\documentclass{llncs_mod}   %% for arxiv

\usepackage[T1]{fontenc}
\usepackage[utf8]{inputenc}

\usepackage{graphicx}
\usepackage[shortlabels]{enumitem}
\usepackage{amsmath}
\usepackage{amssymb}
\usepackage{amsfonts}
\usepackage{mathtools}

\usepackage[absolute]{textpos}

\usepackage{hyperref}  %% for arxiv
\hypersetup{pdfborder={0 0 0},colorlinks=true,urlcolor=blue,linkcolor=blue,citecolor=blue,bookmarks=true}
\hypersetup{pdftitle={Predicting Graph Structure via Adapted Flux Balance Analysis}}
\hypersetup{pdfauthor={Sevvandi Kandanaarachchi, Ziqi Xu, Stefan Westerlund, Conrad Sanderson}}
\usepackage[british]{babel}
\usepackage{hyphenat}
\usepackage{booktabs}
\usepackage[misc]{ifsym}

\usepackage{bm}
\usepackage{multirow}

\usepackage{fancyvrb}  % for Verbatim environment
\usepackage{newtxtext}  % replacement for times package
\usepackage{newtxmath}  % replacement for math font
\usepackage{inconsolata}  % changes the default typewriter font
\usepackage{dsfont}

\newcommand\bluesout{\bgroup\markoverwith{\textcolor{blue}{\rule[0.5ex]{2pt}{0.4pt}}}\ULon}

\begin{document}

\def\spacingset#1{\renewcommand{\baselinestretch}%
{#1}\small\normalsize} \spacingset{1}

%%%%%%%%%%%%%%%%%%%%%%%%%%%%%%%%%%%%%%%%%%%%%%%%%%%%%%%%%%%%%%%%%%%%%%%%%%%%%%

\titlerunning{~}
\authorrunning{~}

\title
  {
  \scalebox{0.89}{Predicting Graph Structure via Adapted Flux Balance Analysis}
  }

\author
  {
  Sevvandi Kandanaarachchi\textsuperscript{{\tiny~}$\dagger$},
  Ziqi Xu\textsuperscript{{\tiny~}$\ddagger$},
  Stefan Westerlund\textsuperscript{{\tiny~}$\dagger$},
  Conrad Sanderson\textsuperscript{{\tiny~}$\dagger\diamond$}
  }

\institute
  {
  \textsuperscript{$\dagger$}{\tiny~}\textit{CSIRO, Australia;}~
  \textsuperscript{$\ddagger$}{\tiny~}\textit{RMIT University, Australia;}~
  \textsuperscript{$\diamond$}{\tiny~}\textit{Griffith University, Australia}
  }

\maketitle

\begin{abstract}
Many dynamic processes such as telecommunication and transport networks
can be described through discrete time series of graphs.
Modelling the dynamics of such time series enables
prediction of graph structure at future time steps,
which can be used in applications such as detection of anomalies.
Existing approaches for graph prediction have limitations
such as assuming that the vertices do not to change between consecutive graphs.
To address this, we propose to exploit time series prediction methods
in combination with an adapted form of flux balance analysis (FBA),
a~linear programming method originating from biochemistry.
FBA is adapted to incorporate various constraints applicable to the scenario of growing graphs.
Empirical evaluations on synthetic datasets (constructed via Preferential Attachment model)
and real datasets (UCI Message, HePH, Facebook, Bitcoin)
demonstrate the efficacy of the proposed approach.
\end{abstract}

\begin{textblock}{9.1}(3.5,14.0)
\hrule
\vspace{1ex}
\noindent
\scriptsize
\textbf{{$^\ast$}~Published in:} Lecture Notes in Computer Science (LNCS), Vol.~16370, pp.~351–363, 2026.
\end{textblock}

\begin{textblock}{9.1}(3.5,14.3)
\noindent
\scriptsize
{\tt ~} DOI:~\href{https://doi.org/10.1007/978-981-95-4969-6_27}{10.1007/978-981-95-4969-6\_27}
\end{textblock}

\renewcommand{\baselinestretch}{1.05}\small\normalsize

\section{Introduction}
\label{sec:introduction}
% \vspace{-1ex}

Dynamic processes such as transport, electricity, telecommunication, and social networks
can be represented through an ordered sequence of graphs.
Such sequences, also known as dynamic graphs,
describe how aspects of graphs evolve over time,
such as the addition and deletion of vertices (nodes) and edges (links between nodes)~\cite{Kazemi2020}.
Modelling the observed dynamics in such sequences can be used for predicting graphs at future time steps.
This in turn can facilitate various applications,
such as the detection of anomalies (differences between predicted and observed graphs),
thereby allowing active response to events of interest
(eg., network overloads, cyber attacks, car accidents)~\cite{Ekle_2024,Kandanaarachchi_2024}.

More formally, given an ordered time sequence of observed graphs
{\small $\left\{ \mathcal{G}_1, \mathcal{G}_2, \ldots, \mathcal{G}_T \right\}$}, 
we aim to predict the graph structure (vertices and edges) for a future time step {\small $T+h$}.
Existing approaches related to graph prediction have notable limitations.
For example, in \textit{link prediction},
where the task is to predict the presence of links between vertices,
the vertices are assumed not to change between consecutive graphs~\cite{Kumar2020}.
In \textit{network time series forecasting}, 
node attributes are predicted while the structure of the network is assumed to be fixed and known~\cite{KIM2024971}.
To our knowledge, the task of graph prediction involving
changes in the number of vertices and edges has not been studied before.

In this work we explore the feasibility of combining time series prediction
with Flux Balance Analysis~(FBA) \cite{Orth2010,Sahu_2021}
for the task of predicting graph structures, 
% for the task of predicting graph structures%
% \footnote
%   {
%   Associated source code in Python and R is available at \url{TODO}
%   }%
% , 
where vertices and edges are added at each time step (ie.~growing graphs).
FBA is a mathematical approach widely used in biochemistry to reconstruct metabolic networks
from partial information by using linear and mixed-integer programming.
By solving an optimisation problem maximising a biomass function subject to a large number of constraints,
FBA~arrives at the flux solution describing a network of chemical reactions.
We adapt FBA to graph prediction by considering the degree of each vertex
instead of the flux of chemical reactions.
To predict a graph given a sequence of previous graphs,
we incorporate the degree prediction of each vertex in the constraints.

We continue the paper as follows.
The proposed adaptation of FBA to graph prediction is described in Section~\ref{sec:graph_prediction}.
The efficacy of the proposed approach is evaluated on synthetic and real-world datasets in Section~\ref{sec:experiments}.
The main findings and future avenues of research are summarised in Section~\ref{sec:conclusion}.

\vspace{-1ex}
\section{Graph Prediction}
\label{sec:graph_prediction}
\vspace{-1ex}

We consider an ordered sequence (time series) of undirected and unweighted graphs without loops
{\small $\mathcal{G}_t = \left(V_t, E_t\right)$}, 
where
{\small $t$} indicates a discrete time index,
{\small $V_t$} denotes the set of vertices,
and {\small $E_t$} denotes the set of edges.
Let {\small $V_t = \{v_1, \ldots, v_{n_t}\}$} denote the set of vertices in graph {\small $\mathcal{G}_t$},
with the number of vertices denoted by {\small $n_t = \vert V_t \vert$}.
Let {\small $e_{ij,t}$} indicate the presence of an edge between vertices {\small $i$} and~{\small $j$} at time {\small $t$},
and let {\small $m_t = \vert E_t \vert$} denote the number of edges in~{\small $\mathcal{G}_t$}.
The set of edges in {\small $\mathcal{G}_t$} is hence {\small $E_t = \{e_{ij,t}\}_{i, j\in V_{t}}$}.  
Let the degree of vertex {\small $v_i$} (number of edges connected to the vertex)
at time {\small $t$} be denoted as~{\small $d_{i,t}$}.
In~an unweighted graph setting, {\small $d_{i,t} = \sum_j e_{ij,t}$}.
Hat notation is employed to indicate the predicted versions of variables;
for example, {\small $\widehat{\mathcal{G}}_{t}$} indicates the predicted version of {\small ${\mathcal{G}}_{t}$}.

We consider the scenario of growing graphs, where {\small $V_t \subseteq V_{t+1}$}.
We aim to obtain the prediction {\small $\widehat{\mathcal{G}}_{T+h} = ( \widehat{V}_{T+h}, \widehat{E}_{T+h} )$},
describing the structure of the graph at time {\small $T+h$},
which includes new vertices and new edges. 
We propose to generate the prediction {\small $\widehat{\mathcal{G}}_{T+h}$} via two main steps:
\textbf{(i)} predict the number of vertices {\small $\widehat{n}_{T+h}$} and their corresponding degree distributions,
\textbf{(ii)} predict the edges using the predicted degree distributions, taking into account new vertices.
The above steps are expanded as follows.

From the given graph time series {\small $\{\mathcal{G}_t\}_{t=1}^T$}
we first extract the corresponding {\small $\{n_t\}_{t=1}^T$} and~{\small $\{d_{i,t}\}_{t=t_{0,i}}^T$} time series,
where {\small $t_{0,i}$} indicates the time step when the $i$-th vertex first appeared.
For clarity, {\small $\{n_t\}_{t=1}^T$} represents the time series of the number of vertices,
while {\small $\{d_{i,t}\}_{t=t_{0,i}}^T$} represents the time series of the degree of vertex $i$.
We note that for each vertex $i$ there is a separate time series.

Employing ARIMA time series modelling with automatic parameter selection~\cite{HyndmanBook2013},
{\small $\{n_t\}_{t=1}^T$} is used for predicting {\small $\widehat{n}_{T+h}$}.
The number of new vertices is {\small $\widehat{n}_{T+h} - n_{T}$}.
The time series {\small $\{d_{i,t}\}_{t=t_{0,i}}^T$} is used for predicting {\small $\widehat{d}_{i, T+h}$} for vertices that already exist
(ie.~vertices that are in both {\small $\mathcal{G}_{T}$} and {\small $\widehat{\mathcal{G}}_{T+h}$}).
Let us define the term \textit{$t$-new vertices} as vertices that are in $\mathcal{G}_{t}$ but are not in $\mathcal{G}_{t-1}$,
which is applicable to both existing graphs (ie.,~training data) and predicted graphs.
The predicted degree {\small $\widehat{d}_{i,T+h}$} for $(T+h)$-new vertices $v_i$ in {\small $\widehat{\mathcal{G}}_{T+h}$}
is taken as the average degree of $t$-new vertices added in the previous time steps for {\small $1 < t \leq T$}.
Distribution of edges among the vertices must satisfy the predicted degree distribution,
which we treat as an optimisation problem.
We propose to use Flux Balance Analysis~\cite{Orth2010,Sahu_2021} to allocate the edges
and hence obtain the predicted graph~{\small $\widehat{\mathcal{G}}_{T+h}$}.

%
%
%

%\vspace{-1ex}
\subsection{Flux Balance Analysis (FBA)}
\label{sec:fba_summary}
%\vspace{-1ex}

FBA was originally designed to reconstruct metabolic networks,
using linear and integer programming.
In a traditional application of FBA,
a large pool of chemical reactions is used as the input.
By mathematically representing chemical reactions and compounds using stoichiometric coefficients,
a set of constraints is obtained.
The solution space, defined by the set of constraints describing the potential network,
is denoted by the matrix equation {\small $\bm{S}\bm{x} = 0$},
with associated inequalities
\mbox{\small $a_i \leq x_i \leq b_i$} for \mbox{\small $i \in \{1, \ldots, q \}$},
where {\small $\bm{x}$} denotes the vector of fluxes to be determined,
and {\small $\bm{S}$} denotes the constraint matrix with size {\small $p~\times~q$},
where $p$ represents the number of constraints and $q$ represents the number of variables $x_i$.
In typical applications of FBA, $p < q$,
meaning we have an under-determined system of linear equations,
which in turn leads to multiple possible solutions.
The solution deemed as optimal is selected by maximising a growth function denoted by $Z = \sum_i c_i x_i$,
where $c_i$ are optimisation coefficients \cite{Orth2010,Sahu_2021}.

%
%
%

%\vspace{-1ex}
\subsection{Adapting FBA to Graph Prediction}
\label{sec:lptographs}
%\vspace{-1ex}

FBA considers the following optimisation problem:
\begin{align}
  \small
  \max  \sum\nolimits_{i}  c_i x_i ~~~ \text{with constraints} ~~ \bm{S}\bm{x} = 0 ~~ \text{and} ~~ \bm{a} \leq \bm{x} \leq \bm{b}
  \label{eq:optimiseFBA}
 \end{align}

\noindent
Our task is to predict {\small $\mathcal{G}_{T+h}$},
denoted by {\small $\widehat{\mathcal{G}}_{T+h}$}
from the graph time series {\small $\left\{ \mathcal{G}_t\right\}_{t = 1}^T$}.
To this end, we adapt FBA by considering a tailored version of the optimisation problem:
\begin{align}
  \small
  \max  \sum\nolimits_{e_{ij} \in \mathcal{G}_{T}^H}  \xi_{ij}\widehat{e}_{ij, T+h} ~~~ \text{with constraints}~~ \bm{0} \leq   \bm{S}\bm{u}   \leq f(\bm{d})
  \label{eq:optimiseourversion}
 \end{align}

\noindent
where {\small $\mathcal{G}^{H}_T$} is a hypothetical graph,
$\xi_{ij}$ denotes the optimisation coefficients
for possible edges {\small $\widehat{e}_{ij, T+h} \in \{0, 1\}$} in~{\small $\widehat{\mathcal{G}}_{T+h}$},
$\bm{S}$ denotes the constraint matrix,
{\small $\bm{u}$} denotes an $m$ dimensional binary vector of edges, 
and {\small $f(\bm{d})$} denotes an {\small $\widehat{n}_{T+h}$} dimensional vector of degree predictions.

Each of the above components is described in the following subsections.
In order to predict {\small $\widehat{\mathcal{G}}_{T+h}$}
we consider a hypothetical graph {\small $\mathcal{G}_{T}^H$}
based on the last seen graph {\small $\mathcal{G}_T$}
such that {\small $\mathcal{G}_T \subset \mathcal{G}^{H}_T$}.
The constraint matrix $\bm{S}$ in Eqn.~\eqref{eq:optimiseourversion} is the incidence matrix of the hypothetical graph~{\small $\mathcal{G}^H_T$};
this is covered in Section \ref{sec:incidence_matrix}.
The function {\small $f\left(\bm{d}\right)$} appearing in the constraints
provides the upper bounds of the vertex degrees in {\small $\widehat{\mathcal{G}}_{T+h}$}.
The degree upper bounds $f\left(\bm{d}\right)$ are obtained
by time series prediction of degrees in {\small$\left\{ \mathcal{G}_t\right\}_{t = 1}^T$};
this is covered in Section \ref{sec:degreetimeseries}.
Finally, the optimisation coefficients $\xi_{ij}$ are explained in Section~\ref{sec:optimisation_coefficients}.
The optimisation problem is further refined with additional constraints in Section~\ref{sec:optimisation_problem}.
In Section~\ref{sec:graphdist} we briefly explore the predictive distribution of graphs.

\newpage

%\vspace{-1ex}
\subsection{Hypothetical Graph $\mathcal{G}^{H}_T$ and the Constraint Matrix $S$}
\label{sec:incidence_matrix}
%\vspace{-1ex}

We consider the graph {\small $\mathcal{G}_{T}$}
and construct a hypothetical graph~{\small $\mathcal{G}^{H}_{T}$}
by adding \textbf{(i)}~new vertices and \textbf{(ii)}~new edges to {\small $\mathcal{G}_{T}$}
as detailed below.
We consider the incidence matrix {\small $\bm{S}$} of {\small $\mathcal{G}^{H}_{T}$} as our constraint matrix in Eqn.~\eqref{eq:optimiseourversion}. 

\textbf{Adding new vertices.}
Recall that $n_t$ is the number of vertices in $\mathcal{G}_t$.
We consider the time series {\small$\{n_t\}_{t = 1}^T$} and predict {\small$ \widehat{n}_{T+h}$} via ARIMA modelling.
If the difference {\small$ \widehat{n}_{T+h} - n_T$} is positive,
{\small$ \widehat{n}_{T+h} - n_T$} is the estimated number of new vertices.
Thus the first step in constructing the hypothetical graph is adding {\small $\widehat{n}_{\text{new}} =  \max(\widehat{n}_{T+h} - n_T\, , 0 )$} vertices to~{\small $\mathcal{G}_T$}.

\textbf{Adding new edges.}
New edges can occur in 3 ways:
\textbf{(i)}~when both vertices exist in~{\small $\mathcal{G}_T$},
\textbf{(ii)}~when one vertex exists in~{\small $\mathcal{G}_T$} and the other one is new,
\textbf{(iii)}~when both vertices are new.

For case \textbf{(i)}, new edges are added between existing vertices in {\small $\mathcal{G}_T$} if they share a neighbour,
following the homophily principle~\cite{Khanam_2023,Kim_2017}.
For case \textbf{(ii)}, instead of adding edges from the new vertices to all existing vertices, 
each new vertex is connected to only the $k$-most popular vertices, resulting in the addition of {\small $k~\times~\widehat{n}_{\text{new}}$} edges, 
where the parameter~$k$ is used to control the complexity of the approach (we use a default value of $k = 10$ in the experiments).
The dimensionality of the optimisation problem in Eqn.~\eqref{eq:optimiseourversion}
is equal to the number of edges in {\small $\mathcal{G}_T^H$}.
Adding $k$ new edges to each new vertex hence increases the dimensionality  by {\small $k~\times~\widehat{n}_{\text{new}}$}.
If the limitation of $k$ new edges is not used,
the dimensionality of the optimisation problem increases by {\small $n_T~\times~\widehat{n}_{\text{new}}$},
which can be prohibitively large.

We do not consider case \textbf{(iii)}, as in the short term (eg. {\small $h = 1$})
we assume that each new vertex is added to the graph independently of other vertices being added at the same time;
for example in social networks a new member is more likely to connect to existing members as they unaware of other new members that are joining the network at the same time.

By adding new vertices and new edges to {\small $\mathcal{G}_T$} as described above,
we obtain the hypothetical graph {\small $\mathcal{G}_T^H$}.
The incidence matrix $\bm{S}$ of {\small $\mathcal{G}_T^H$} is the constraint matrix used in Eqn.~\eqref{eq:optimiseourversion}.
Recall that the incidence matrix of a given graph {\small $\mathcal{G}$} with $n$ vertices and $m$ edges has {$n$} rows and {$m$} columns;
the {\small $ij$}-th entry of the matrix is equal to~1 if vertex {\small $v_i$} is incident with edge {\small $e_j$}.
The incidence matrix $\bm{S}$ has {\small $\max( n_T,  \widehat{n}_{T+h})$} rows and $m$ columns,
where $m$ is the number of edges in {\small $\mathcal{G}_T^H$}.

%
%
%

%\vspace{-1ex}
\subsection{Constraint bounds $f(\bm{d})$}
\label{sec:degreetimeseries}
%\vspace{-1ex}

The vector $f(\bm{d})$ in Eqn.~\eqref{eq:optimiseourversion} consists of the predicted upper bound degrees of vertices in {\small $\widehat{\mathcal{G}}_{T+h}$}.
Recall that the degree of a vertex $v_i$ in graph {\small $\mathcal{G}_t$} is given by {\small $d_{i,t} = \sum_j e_{ij,t}$}.
We~consider the vertices in {\small $\widehat{\mathcal{G}}_{T+h}$} to be the same as those in the hypothetical graph {\small ${\mathcal{G}}_{T}^H$}.
The hypothetical graph {\small ${\mathcal{G}}_{T}^H$} has two types of vertices:
\textbf{(i)}~existing vertices in {\small ${\mathcal{G}}_{T}$},
and \textbf{(ii)}~new vertices in {\small ${\mathcal{G}}_{T}^H$}.
For existing vertices $v_i$ in {\small ${\mathcal{G}}_{T}$} we use the time series {\small $\{d_{i,t} \}_{t = t_0}^T$}
where $t_0$ denotes the first time at which vertex $v_i$ appears in {\small $\{ \mathcal{G}_t \}_{t=1}^T$}.
Using ARIMA modelling on this time series we predict {\small $\widehat{d}_{i, T+h}$} for each vertex $v_i$ in {\small ${\mathcal{G}}_{T}$}.
The ARIMA implementation we use provides the predicted degree distribution in addition to the mean estimate {\small $\widehat{d}_{i, T+h}$}. 
Using this predictive distribution we can for example select the 80-th percentile of the predicted degree distribution for a given vertex.
This upper bound is denoted by $f$.
We use a user-defined percentile to obtain the upper bound {\small $f(\widehat{d}_{i, T+h})$}.
Thus, for existing vertices in {\small ${\mathcal{G}}_{T}$} we obtain the degree upper bound.

We predict the degrees of new vertices in {\small ${\mathcal{G}}_{T}^H$} in a different way.
Consider the graph time series {\small $\{ \mathcal{G}_t \}_{t=1}^T$}.
We refer to vertices that are in {\small $ \mathcal{G}_{t} $} but not in {\small $\mathcal{G}_{t-1}$} as \mbox{$t$-new vertices}.
Let the set of degrees of the $t$-new vertices in {\small $ \mathcal{G}_{t} $} be~given~by \mbox{\small $D_{t_\text{new}} = \{ d_{i, t}: i \in t\text{-new vertices} \}$}.
For each graph {\small $\mathcal{G}_{t}$} we observe {\small $D_{t_\text{new}}$}.
Let {\small $D_{T} = \bigcup_{t \leq T} D_{t_\text{new}}$} denote the union of all degree values in {\small $D_{t_\text{new}}$} for {\small $t \leq T$},
which results in {\small $D_{T}$} containing the degrees of $t$-new vertices for $t \leq T$.
We compute the mean of the $t$-new vertices for $t \leq T$ denoted by {\small $\text{mean}(D_T)$},
and assign {\small $f(\widehat{d}_{i,T+h}) = \text{mean}(D_T)$} for vertices $v_i$ that are new in  {\small ${\mathcal{G}}_{T}^H$}.
Thus, we have degree upper bounds for both existing vertices in {\small${\mathcal{G}}_{T}$} and new vertices in {\small ${\mathcal{G}}_{T}^H$}.

%
%
%

%\vspace{-1ex}
\subsection{Optimisation Coefficients $\xi_{ij}$}
\label{sec:optimisation_coefficients}
%\vspace{-1ex}

The optimisation coefficients $\xi_{ij}$ in Eqn.~\eqref{eq:optimiseourversion}
are computed using the last seen graph {\small ${\mathcal{G}}_{T}$}
and the hypothetical graph {\small${\mathcal{G}}_{T}^H$} via:

\vspace{0.5ex}
\noindent
\begin{equation}
  \small
  \xi_{ij} =
  \begin{cases}
   \mathds{1}_{e_{ij, T} \in \mathcal{G}_T}  &  \text{for existing edges } e_{ij}   \text{ in } \mathcal{G}_{T} \\
   \alpha &  \text{for new edges in $\mathcal{G}^{H}_{T}$}  \\
\end{cases}
  \label{eq:simplecoeffs}
\end{equation}

\noindent
where $\mathds{1}$ denotes the indicator function, 
$\alpha$ indicates a fixed prior likelihood of the presence of edges for new vertices.
The reasoning is as follows.
We assign 1 for every edge in graph {\small $\mathcal{G}_T$}, which are existing edges and are likely to persist.
In a growing graph scenario, we expect to have new edges at time {\small $T+h$}.
The possible new edges are accounted for in {\small $\mathcal{G}^{H}_{T}$}.
These new edges can be between existing vertices or between existing and new vertices.
For these new edges we assign $\alpha$, reflecting a low certainty of their presence.

%
%
%
 
%\vspace{-1ex}
\subsection{Refining the Optimisation Problem via Further Constraints}
\label{sec:optimisation_problem}
%\vspace{-1ex}

Recall that the constraint matrix $\bm{S}$ in Eqn.~\eqref{eq:optimiseourversion} 
is the incidence matrix of the hypothetical graph {\small $\mathcal{G}_{T}^H$} 
and the vector {\small $f(\bm{d})$} contains the predicted degree upper bounds.
We refine the optimisation problem by adding an additional constraint,
which bounds the total number of edges in {\small $\widehat{\mathcal{G}}_{T+h}$}.
The reasoning is that each constraint bounds the degree of a single vertex,
which may result in the overall addition of many superfluous edges to the graph.
Furthermore, individual vertices have more randomness in their degree evolution compared with the total number of edges.
The total number of edges from one time step to the next is generally more stable than the degree fluctuation of individual vertices from one time step to the next.
For example, consider a member of a social media network connecting with other members;
on a given day the member might make 5 new connections,
but the next day not make any connections as they did not use the network.
As such, instead of considering individual vertices, it is more robust to consider the growth of the total edges in the graph.

Let {\small $|E_t|$} denote the number of edges in graph {\small $\mathcal{G}_t$}.
We consider the time series {\small$ \left \{ |E_t| \right\}_{t = 1}^T$} and as before predict {\small$ {|\widehat{E}_{T+h}|}$} via ARIMA modelling.
This is the predicted total edges of the graph.
As in Section~\ref{sec:degreetimeseries} we can take the upper bound percentile of the predicted distribution
and obtain {\small $f \left( \vert \widehat{E}_{T+h} \vert \right)$}.
This gives us:

\noindent
\begin{equation}
  \sum\nolimits_{e_{ij} \in \mathcal{G}_{T}^H} \widehat{e}_{ij} ~ \leq ~ f \left( \vert \widehat{E}_{T+h} \vert \right)
\end{equation}

\noindent
The above constraint is incorporated to the constraint matrix $\bm{S}$ by adding a new row of ones as the last row of $\bm{S}$.
The vector {\small $f(\bm{d})$} is concatenated with {\small $f ( \vert \widehat{E}_{T+h} \vert )$} as its new last element. 

%
%
%

%\vspace{-1ex}
\subsection{Prediction Distribution of Graphs}
\label{sec:graphdist}
%\vspace{-0.5ex}

By solving the optimisation problem described in Section~\ref{sec:optimisation_problem}
we obtain a possible graph at time {\small $T+h$}, denoted by {\small $\widehat{\mathcal{G}}_{T+h}$}.
This graph is a prediction and hence can be treated as an instance drawn from a prediction distribution,
obtained by making certain choices on {\small $\widehat{n}_{T+h}$} and {\small $f(\bm{d})$}.

\begin{figure}[!t]
  \centering
  \includegraphics[width=0.5\textwidth]{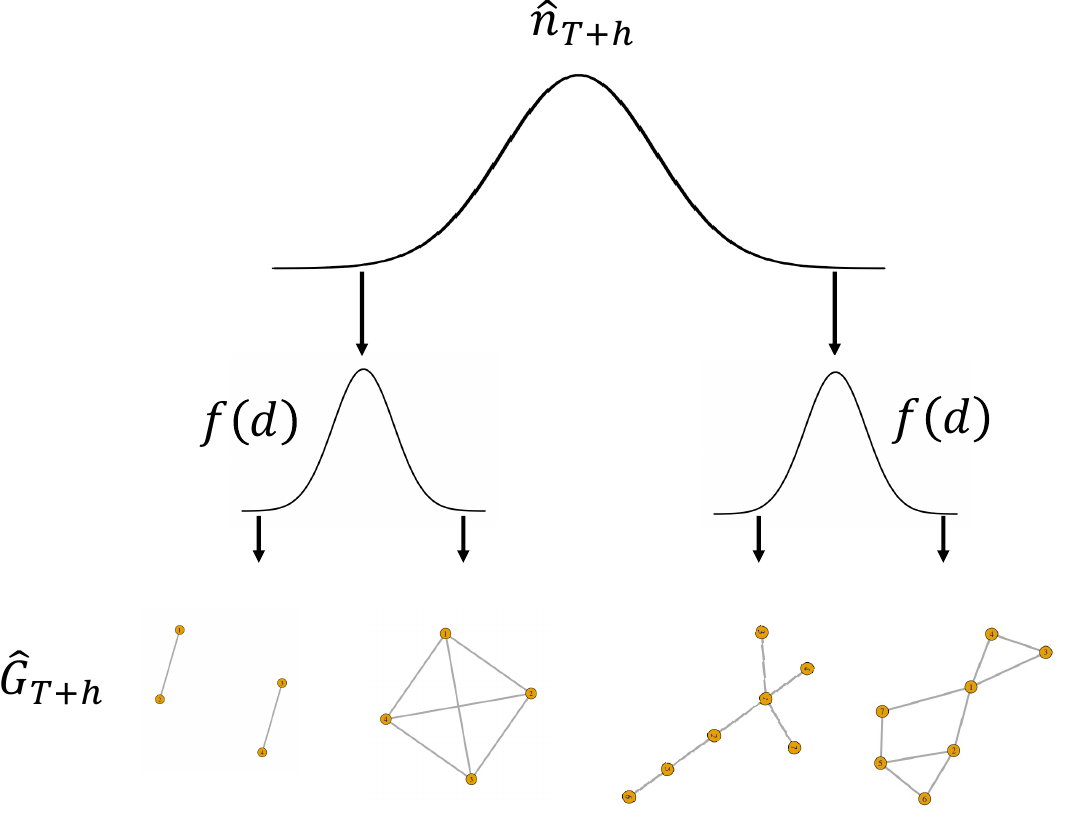}
  \caption
    {
    A conceptual illustration of the prediction graph distribution in terms of $\gamma$ and $u$.
    The parameter $\gamma$ gives various $\widehat{n}_{T+h}$ values.
    The two graphs on bottom left each have 4 vertices, corresponding to a single $\gamma$ (or $\widehat{n}_{T+h}$) value,
    but have differing number of edges corresponding to various $u$ values.
    Similarly, the two graphs on the bottom right have 7 vertices corresponding to a higher value of $\gamma$
    with the rightmost graph having more edges due to a higher $u$.
    }
  \label{fig:networkdistribution}
  \vspace{1ex}
  \hrule
  \vspace{1ex}
\end{figure}

Figure~\ref{fig:networkdistribution} illustrates the prediction graph distribution
where {\small $\widehat{\mathcal{G}}_{T+h}$} is a function of prediction vertices {\small $\widehat{n}_{T+h}$} and {\small $f(\bm{d})$}.
If we explore the predictions for various values of {\small $\widehat{n}_{T+h}$} and various upper bounds $u$, we get various graphs.
Let $\gamma$ denote the quantile used to obtain {\small $\widehat{n}_{T+h}$}
and $u$ denote the quantiles in $f(\bm{d})$.
Then the graph {\small $\widehat{\mathcal{G}}_{T+h}$} can be considered
as belonging to a graph distribution {$\mathfrak{G}$} having parameters $\gamma$ and $u$:
{\small $\widehat{\mathcal{G}}_{T+h} \sim \mathfrak{G}\left(\gamma, u \right)$}.
While these are explicit parameters, the coefficient scheme used to obtain $\xi_{ij}$,
as well as the methods used for predicting individual time series and their hyperparameters
can also be considered as part of a wider parameter pool.

%
%
%

%\vspace{-1ex}
\section{Empirical Evaluation}
\label{sec:experiments}
\vspace{-1ex}

We investigate the performance of the proposed FBA-based graph prediction approach on synthetic and real graphs.
For all experiments we consider a time series of graphs {\small $\{ \mathcal{G}_t\}_{t=1}^T$}
and predict graphs at future time steps for {\small $h \in \{1, 2, 3, 4, 5 \}$}
using the proposed approach.
Based on preliminary evaluations, all experiments use {$\alpha = 1 \times 10^{-3}$} in Eqn.~(\ref{eq:simplecoeffs}),
which reflects low certainty of the presence of new edges.
Furthermore,  $k = 10$, in order to keep the computational complexity relatively low (see Section~\ref{sec:incidence_matrix}).

There are well established evaluation metrics for time series prediction of real valued quantities,
such as the Mean Absolute Percentage Error~\cite{Myttenaere_2016}.
However, for the task of graph prediction from a time series of graphs, evaluation becomes more challenging.
One option is to check if the predicted graph~{\small $\widehat{\mathcal{G}}_{T+h}$} is isomorphic to the actual graph~{\small ${\mathcal{G}}_{T+h}$}.
However, this is computationally expensive~\cite{graphiso}.
Another option is via graph neural networks and graph embeddings~\cite{Liu_2025,ma2021deep,Souid_2024},
though these come with their own issues, including acting as black boxes which reduce interpretability~\cite{Arrieta_2020,Sanderson_2023}.
With the aim of having readily interpretable and low complexity evaluation measures,
we have elected use the following straightforward errors as proxies for graph similarity,
with lower values indicating better performance:

\noindent
\begin{itemize}[{$\bullet$},leftmargin=*]
\itemsep=0.5ex

\item
\textbf{vertex error}, defined as {\small $\vert \widehat{n}_{T+h} - n_{T+h} \vert ~/~ {n_{T+h}}$}
which provides the absolute error ratio of the number of vertices in the predicted graph compared to the actual graph;

\item
\textbf{edge error}, defined as {\small $\vert \widehat{m}_{T+h}  -   m_{T+h} \vert / {m_{T+h}}$}
which provides the absolute error ratio of the number of edges in the predicted graph compared to the actual graph.

\end{itemize}
\vspace{-1ex}

\vspace{-1ex}
\subsection{Synthetic Graphs}
\label{sec:synthetic}
\vspace{-0.5ex}

We generate a growing sequence of random graphs {\small $\left \{\mathcal{G}_t \right \}_{t=1}^T$}
following the Preferential Attachment (PA) model~\cite{barabasi1999}.
The PA model considers vertices connecting to more connected vertices with a higher probability.
The linear PA model specifies the probability~{\small $\Pi(k)$} of a new vertex connecting to vertex~$i$ with degree~$k_i$
to be {\small $\Pi(k_i) = \frac{k_i}{\sum_i k_i}$}.
At each time step, the PA model adds a new vertex with $s$ edges.
After~$t$ time steps,
the network has $n_t = t + s_0$ vertices and $m_t = s_0 + ts$ edges,
where at time $t = 0$ the network has $s_0$ nodes and edges. 

In the first experiment,
we consider a sequence of 20 PA graphs with {\small $s = 10$}
and $n_t$ vertices in~{\small $\mathcal{G}_t$},
with $n_t$ randomly sampled from {\small $\{45 + 5t, 46 + 5t, \ldots, 49 + 5t \}$}.
This sampling scheme ensures that $n_t$ is non-deterministic while it increases with $t$,
aiming to mimic more realistic scenarios.
Using the first 15 graphs, we predict the next 5 graphs,
i.e.,~we set {\small $T = 15$} and predict {\small $\widehat{\mathcal{G}}_{T+h}$} for {\small $h \in \{1, 2, 3, 4, 5 \}$}.
The second experiment is similar to the first experiment,
with the difference that after each graph growth we delete $r$ edges, where $r$ is randomly selected from {\small $\{5, \ldots, 10\}$}.

Each experiment is executed 10 times to take into account the randomness in graph generation via the PA model.
The results are presented in Table~\ref{tab:synthetic_data}
in terms of mean vertex error and mean edge error over the 10 instances.
To place the obtained errors of the proposed approach into context, 
we also provide the corresponding errors when the last seen graph {\small $\mathcal{G}_T$} is used as the predicted graph.

The results demonstrate that the proposed FBA-based approach produces graphs which have considerably lower errors compared to using the last seen graph.
Reductions in the vertex error (compared to the last seen graph) range from approximately {\small $70\%$} to~{\small $94\%$},
while corresponding reductions in the edge error range from about {\small $35\%$} to~{\small $79\%$}.
For both error types, the largest reductions tend to occur for {\small $h=5$} (ie.~the furthest step from the last seen graph),
indicating that the prediction mechanism is working.

\begin{table}[!t]
%\vspace{-1ex}
%\hrule
%\vspace{1ex}
\caption
  {
  Results for experiments on random graphs generated using the Preferential Attachment model~\cite{barabasi1999}.
  Given a sequence of 20 graphs, the first $T=15$ graphs are used for training the proposed approach,
  followed by predicting graphs at time step $T+h$, where $h \in \{1, 2, 3, 4, 5 \}$.
  The predicted graphs are compared to the actual graphs, using vertex error and edge error as measurements.
  The errors are reported as averages over 10 instances of each experiment, to account for randomness in the generated graphs.
  For context, corresponding errors are also reported when the last seen graph at $T=15$ is used as the predicted graph.
  The reduction in error is denoted as a percentage, where a given error produced by the proposed approach is compared to the corresponding error produced by the last seen graph.
  \label{tab:synthetic_data}
  }
%\fontsize{8.4}{8.6}\selectfont
\fontsize{8.6}{8.8}\selectfont
\begin{minipage}{1\textwidth}
\centering
\begin{minipage}{0.47\textwidth}
\centering
% \scriptsize
\textbf{Experiment 1}\\
\vspace{0.5ex}
\setlength\tabcolsep{3pt}
\begin{tabular}{cccccccc}
  \toprule
  $h$              & {\bf method} & {\bf vertex error} & {\bf edge error} \\ \midrule
  \multirow{2}*{1} & last seen & \tt $~35.4 {\hspace{0.5pt}\times\hspace{0.5pt}} 10^{-3}$ & \tt $~37.0 {\hspace{0.5pt}\times\hspace{0.5pt}} 10^{-3}$ \\
  ~                & proposed  & \tt $~10.3 {\hspace{0.5pt}\times\hspace{0.5pt}} 10^{-3}$ & \tt $~24.1 {\hspace{0.5pt}\times\hspace{0.5pt}} 10^{-3}$ \\ 
  & {\scriptsize (reduction)}  & \tt $~~~~~70.9\%$                                        & \tt $~~~~~34.9\%$                                        \\ \midrule
  \multirow{2}*{2} & last seen & \tt $~72.0 {\hspace{0.5pt}\times\hspace{0.5pt}} 10^{-3}$ & \tt $~75.2 {\hspace{0.5pt}\times\hspace{0.5pt}} 10^{-3}$ \\
  ~                & proposed  & \tt $~11.4 {\hspace{0.5pt}\times\hspace{0.5pt}} 10^{-3}$ & \tt $~30.4 {\hspace{0.5pt}\times\hspace{0.5pt}} 10^{-3}$ \\ 
  & {\scriptsize (reduction)}  & \tt $~~~~~84.2\%$                                        & \tt $~~~~~59.6\%$                                        \\ \midrule
  \multirow{2}*{3} & last seen & \tt $107.3 {\hspace{0.5pt}\times\hspace{0.5pt}} 10^{-3}$ & \tt $111.8 {\hspace{0.5pt}\times\hspace{0.5pt}} 10^{-3}$ \\
  ~                & proposed  & \tt $~10.9 {\hspace{0.5pt}\times\hspace{0.5pt}} 10^{-3}$ & \tt $~31.9 {\hspace{0.5pt}\times\hspace{0.5pt}} 10^{-3}$ \\ 
  & {\scriptsize (reduction)}  & \tt $~~~~~89.8\%$                                        & \tt $~~~~~71.5\%$                                        \\ \midrule
  \multirow{2}*{4} & last seen & \tt $134.5 {\hspace{0.5pt}\times\hspace{0.5pt}} 10^{-3}$ & \tt $140.0 {\hspace{0.5pt}\times\hspace{0.5pt}} 10^{-3}$ \\
  ~                & proposed  & \tt $~~7.8 {\hspace{0.5pt}\times\hspace{0.5pt}} 10^{-3}$ & \tt $~40.7 {\hspace{0.5pt}\times\hspace{0.5pt}} 10^{-3}$ \\ 
  & {\scriptsize (reduction)}  & \tt $~~~~~94.2\%$                                        & \tt $~~~~~70.9\%$                                        \\ \midrule
  \multirow{2}*{5} & last seen & \tt $168.6 {\hspace{0.5pt}\times\hspace{0.5pt}} 10^{-3}$ & \tt $175.2 {\hspace{0.5pt}\times\hspace{0.5pt}} 10^{-3}$ \\
  ~                & proposed  & \tt $~11.6 {\hspace{0.5pt}\times\hspace{0.5pt}} 10^{-3}$ & \tt $~36.3 {\hspace{0.5pt}\times\hspace{0.5pt}} 10^{-3}$ \\
  & {\scriptsize (reduction)}  & \tt $~~~~~93.1\%$                                        & \tt $~~~~~79.3\%$                                        \\ 
  \bottomrule
  \end{tabular}
\end{minipage}
\hfill
\begin{minipage}{0.47\textwidth}
\centering
\textbf{Experiment 2}\\
\vspace{0.5ex}
\setlength\tabcolsep{3pt}
\begin{tabular}{cccccccc}
  \toprule
  $h$              & {\bf method} & {\bf vertex error} & {\bf edge error} \\ \midrule
  \multirow{2}*{1} & last seen & \tt $~40.7 {\hspace{0.5pt}\times\hspace{0.5pt}} 10^{-3}$ & \tt $~42.6 {\hspace{0.5pt}\times\hspace{0.5pt}} 10^{-3}$ \\
  ~                & proposed  & \tt $~10.2 {\hspace{0.5pt}\times\hspace{0.5pt}} 10^{-3}$ & \tt $~18.5 {\hspace{0.5pt}\times\hspace{0.5pt}} 10^{-3}$ \\ 
  & {\scriptsize (reduction)}  & \tt $~~~~~74.9\%$                                        & \tt $~~~~~56.6\%$                                        \\ \midrule
  \multirow{2}*{2} & last seen & \tt $~72.6 {\hspace{0.5pt}\times\hspace{0.5pt}} 10^{-3}$ & \tt $~75.9 {\hspace{0.5pt}\times\hspace{0.5pt}} 10^{-3}$ \\
  ~                & proposed  & \tt $~12.9 {\hspace{0.5pt}\times\hspace{0.5pt}} 10^{-3}$ & \tt $~29.9 {\hspace{0.5pt}\times\hspace{0.5pt}} 10^{-3}$ \\ 
  & {\scriptsize (reduction)}  & \tt $~~~~~82.2\%$                                        & \tt $~~~~~60.6\%$                                        \\ \midrule
  \multirow{2}*{3} & last seen & \tt $108.3 {\hspace{0.5pt}\times\hspace{0.5pt}} 10^{-3}$ & \tt $112.6 {\hspace{0.5pt}\times\hspace{0.5pt}} 10^{-3}$ \\
  ~                & proposed  & \tt $~13.8 {\hspace{0.5pt}\times\hspace{0.5pt}} 10^{-3}$ & \tt $~33.6 {\hspace{0.5pt}\times\hspace{0.5pt}} 10^{-3}$ \\ 
  & {\scriptsize (reduction)}  & \tt $~~~~~87.3\%$                                        & \tt $~~~~~70.2\%$                                        \\ \midrule
  \multirow{2}*{4} & last seen & \tt $136.6 {\hspace{0.5pt}\times\hspace{0.5pt}} 10^{-3}$ & \tt $142.5 {\hspace{0.5pt}\times\hspace{0.5pt}} 10^{-3}$ \\
  ~                & proposed  & \tt $~~9.2 {\hspace{0.5pt}\times\hspace{0.5pt}} 10^{-3}$ & \tt $~39.0 {\hspace{0.5pt}\times\hspace{0.5pt}} 10^{-3}$ \\ 
  & {\scriptsize (reduction)}  & \tt $~~~~~93.3\%$                                        & \tt $~~~~~72.6\%$                                        \\ \midrule
  \multirow{2}*{5} & last seen & \tt $170.0 {\hspace{0.5pt}\times\hspace{0.5pt}} 10^{-3}$ & \tt $176.6 {\hspace{0.5pt}\times\hspace{0.5pt}} 10^{-3}$ \\
  ~                & proposed  & \tt $~10.8 {\hspace{0.5pt}\times\hspace{0.5pt}} 10^{-3}$ & \tt $~36.3 {\hspace{0.5pt}\times\hspace{0.5pt}} 10^{-3}$ \\
  & {\scriptsize (reduction)}  & \tt $~~~~~93.6\%$                                        & \tt $~~~~~79.4\%$                                        \\ 
  \bottomrule
  \end{tabular}
\end{minipage}
\end{minipage}

\end{table}

\vspace{-1ex}
\subsection{Real Graphs}
\vspace{-0.5ex}

We use four datasets containing graphs obtained from real-life data:
%
%\begin{enumerate}[{1.},leftmargin=*]
\begin{enumerate}[{$\bullet$},leftmargin=*]
%\small
\fontsize{9.75}{10.75}\selectfont   %% this provides more control over fontsize compared to \small
\itemsep=0.5ex

\item
\textbf{UCI Message (UCI) dataset},
containing interactions between an online community of students at the University of California Irvine~\cite{Panzarasa2009}.
Each student is represented by a vertex and communications between students are represented by edges.

\item
\textbf{High energy physics citations (HePH) dataset},
containing citations of high energy physics papers published on the arXiv pre-print server~\cite{Leskovec2007}.
Each paper is denoted as a vertex and citations between papers are denoted as edges. 

\item
\textbf{Facebook (FB) dataset},
containing users and their links from the Facebook New Orleans networks~\cite{viswanath-2009-activity}.
Users are depicted as vertices, with an edge present between two users if they are friends.

\item
\textbf{Bitcoin dataset},
where users rate the degree of trust they have in other users~\cite{kumar2016edge},
an important aspect in bitcoin transactions.
We focus on the growth of this rating network.

\end{enumerate}

The datasets are represented via anonymised lists of edge observations in the form {\small $(u, v, t)$},
which denotes an edge between vertices $u$ and $v$ at time $t$.
From this list of time stamped edges,
we construct a sequence of growing graphs by considering edges belonging to expanding time windows.
By expanding time windows we mean that {\small $\mathcal{G}_1$} is constructed from edges $(u, v, t)$ for $t \leq t_1$
and {\small $\mathcal{G}_2$} is constructed from edges $(u, v, t)$ for $t \leq t_2$ where $t_2 > t_1$.
This ensures that all edges in {\small$\mathcal{G}_1$} are also in {\small $\mathcal{G}_2$}.
For example, suppose we want to construct graphs corresponding to  days {\small $\ell \in \{1, \ldots, 5\}$}
and suppose $t_1$, $t_2, \ldots, t_5$ are the times corresponding to the end of each day.  
We consider edges $e_t$ such that {\small $E_\ell = \{e_t | t \leq t_{\ell} \}$}.
We denote by {\small $V_{\ell}$} the vertices that are present in {\small $E_\ell$} and let {\small $\mathcal{G}_\ell = (V_{\ell} , E_{\ell})$}.
Thus, by construction we get a sequence of growing graphs where $E_{\ell} \subset E_{\ell+1}$ and $V_{\ell} \subset V_{\ell+1}$.

For the UCI and FB datasets we consider daily edges to form a graph.
For the HePH dataset we consider bi-weekly edges.
For the bitcoin dataset we use weekly edges.

To standardise computations across the datasets,
we train the proposed approach on graph time series
{\small $\{\mathcal{G}_t \}_{t = {T-14}}^T$} for {\small $T \in \{15, \ldots, 24 \}$}
and predict {\small $\widehat{\mathcal{G}}_{T + h}$} for {\small $h \in \{1, 2, 3, 4, 5 \}$}.
As such, each graph time series has 15 graphs that are used for learning, and is used to predict future graphs.
Predictions were carried out for all combinations of~{\small $T$} and~{\small $h$}.

For clarity, we note that two separate window models are used in the setup for these experiments.
We generated graphs {\small $\{\mathcal{G}_1 ,\ldots ,\mathcal{G}_T, \ldots, \mathcal{G}_{T+h}\}$}
by considering expanding windows on observations $(u,v,t)$ in the edgelist.
A moving window of width 15 is then used on the graph time series {\small $\{\mathcal{G}_t \}_{t = 1}^{29}$}
to train the model and test it on the next 5 graphs.

The results are shown in Table~\ref{tab:real_data} in terms of mean vertex error and mean edge error for each $h$ over all possible values of $T$.
As per the experiments on synthetic graphs, the errors obtained by the proposed approach are placed into context 
by contrasting them with the errors obtained by using the last seen graph {\small $\mathcal{G}_T$} as the predicted graph.
In all cases the errors increase for longer time steps (ie.,~larger values of $h$), which is expected.
However, graphs predicted using the proposed FBA-based approach obtain notably lower errors compared to using the last seen graph.
Across the four datasets, reductions in the vertex error (compared to the last seen graph) range from approximately {\small $52\%$} to~{\small $79\%$},
while reductions in the edge error range from about {\small $20\%$} to~{\small $87\%$}.

The FB dataset appears to be the easiest to predict,
with the reduction in errors staying relatively high ($>70\%$) over the range of $h$ values.
This is expected as homophily plays a big role in social networks~\cite{Aiello_2012}.
In contrast, the Bitcoin dataset appears to be the most difficult to predict;
both error types progressively increase as $h$ increases,
with a decreasing gap between the errors produced by the proposed approach and the last seen graph.
In general, network activity is highly correlated with the exchange rate in Bitcoin networks~\cite{Baumann_2014}.
As we do not use extra covariates such as exchange rate data,
it is more challenging to predict the network multiple time steps ahead.

\begin{table}[!b]
% \hrule
% \vspace{1ex}
\caption
  {
  Results for experiments on graphs obtained from the UCI, HePH, FB and Bitcoin datasets.
  For each dataset, 15 graphs were used for training the proposed approach via {\small $\{\mathcal{G}_t \}_{t = {T-14}}^T$}  
  followed by predicting graphs at time step $T+h$, where $h \in \{1, 2, 3, 4, 5 \}$.
  10~variations of $T$ were used, with {\small $T \in \{15, \ldots, 24 \}$}.
  The predicted graphs are compared to the actual graphs, using vertex error and edge error as measurements.
  The errors are reported as averages over all possible values of~$T$.
  For context, corresponding errors are also reported when the last seen graph at $T$ is used as the predicted graph.
  The reduction in error is denoted as a percentage, where a given error produced by the proposed approach is compared to the corresponding error produced by the last seen graph.
  \label{tab:real_data}
  }
%\fontsize{8.4}{8.6}\selectfont
\fontsize{8.6}{8.8}\selectfont
\begin{minipage}{1\textwidth}
\centering
\begin{minipage}{0.47\textwidth}
\centering
% \scriptsize
\textbf{UCI dataset}\\
\vspace{0.5ex}
\setlength\tabcolsep{3pt}
\begin{tabular}{cccccccc}
  \toprule
  $h$              & {\bf method} & {\bf vertex error} & {\bf edge error} \\ \midrule
  \multirow{2}*{1} & last seen & \tt $~57.00 {\hspace{0.5pt}\times\hspace{0.5pt}} 10^{-3}$ & \tt $~95.12 {\hspace{0.5pt}\times\hspace{0.5pt}} 10^{-3}$ \\
  ~                & proposed  & \tt $~27.07 {\hspace{0.5pt}\times\hspace{0.5pt}} 10^{-3}$ & \tt $~34.36 {\hspace{0.5pt}\times\hspace{0.5pt}} 10^{-3}$ \\ 
  & {\scriptsize (reduction)}  & \tt $~~~~~~52.5\%$                                        & \tt $~~~~~~63.9\%$                                        \\ \midrule
  \multirow{2}*{2} & last seen & \tt $108.17 {\hspace{0.5pt}\times\hspace{0.5pt}} 10^{-3}$ & \tt $177.42 {\hspace{0.5pt}\times\hspace{0.5pt}} 10^{-3}$ \\
  ~                & proposed  & \tt $~35.03 {\hspace{0.5pt}\times\hspace{0.5pt}} 10^{-3}$ & \tt $~60.81 {\hspace{0.5pt}\times\hspace{0.5pt}} 10^{-3}$ \\ 
  & {\scriptsize (reduction)}  & \tt $~~~~~~67.6\%$                                        & \tt $~~~~~~65.7\%$                                        \\ \midrule
  \multirow{2}*{3} & last seen & \tt $149.24 {\hspace{0.5pt}\times\hspace{0.5pt}} 10^{-3}$ & \tt $238.77 {\hspace{0.5pt}\times\hspace{0.5pt}} 10^{-3}$ \\
  ~                & proposed  & \tt $~46.99 {\hspace{0.5pt}\times\hspace{0.5pt}} 10^{-3}$ & \tt $~98.05 {\hspace{0.5pt}\times\hspace{0.5pt}} 10^{-3}$ \\ 
  & {\scriptsize (reduction)}  & \tt $~~~~~~68.5\%$                                        & \tt $~~~~~~58.9\%$                                        \\ \midrule
  \multirow{2}*{4} & last seen & \tt $181.58 {\hspace{0.5pt}\times\hspace{0.5pt}} 10^{-3}$ & \tt $284.55 {\hspace{0.5pt}\times\hspace{0.5pt}} 10^{-3}$ \\
  ~                & proposed  & \tt $~57.89 {\hspace{0.5pt}\times\hspace{0.5pt}} 10^{-3}$ & \tt $125.75 {\hspace{0.5pt}\times\hspace{0.5pt}} 10^{-3}$ \\ 
  & {\scriptsize (reduction)}  & \tt $~~~~~~68.1\%$                                        & \tt $~~~~~~55.8\%$                                        \\ \midrule
  \multirow{2}*{5} & last seen & \tt $209.19 {\hspace{0.5pt}\times\hspace{0.5pt}} 10^{-3}$ & \tt $322.26 {\hspace{0.5pt}\times\hspace{0.5pt}} 10^{-3}$ \\
  ~                & proposed  & \tt $~69.35 {\hspace{0.5pt}\times\hspace{0.5pt}} 10^{-3}$ & \tt $151.30 {\hspace{0.5pt}\times\hspace{0.5pt}} 10^{-3}$ \\
  & {\scriptsize (reduction)}  & \tt $~~~~~~66.8\%$                                        & \tt $~~~~~~53.1\%$                                        \\ 
  \bottomrule
  \end{tabular}
\end{minipage}
\hfill
\begin{minipage}{0.47\textwidth}
\centering
\textbf{FB dataset}\\
\vspace{0.5ex}
\setlength\tabcolsep{3pt}
\begin{tabular}{cccccccc}
  \toprule
  $h$              & {\bf method} & {\bf vertex error} & {\bf edge error} \\ \midrule
  \multirow{2}*{1} & last seen & \tt $~33.45 {\hspace{0.5pt}\times\hspace{0.5pt}} 10^{-3}$ & \tt $~50.48 {\hspace{0.5pt}\times\hspace{0.5pt}} 10^{-3}$ \\
  ~                & proposed  & \tt $~~9.31 {\hspace{0.5pt}\times\hspace{0.5pt}} 10^{-3}$ & \tt $~13.04 {\hspace{0.5pt}\times\hspace{0.5pt}} 10^{-3}$ \\ 
  & {\scriptsize (reduction)}  & \tt $~~~~~~72.2\%$                                        & \tt $~~~~~~74.2\%$                                        \\ \midrule
  \multirow{2}*{2} & last seen & \tt $~64.58 {\hspace{0.5pt}\times\hspace{0.5pt}} 10^{-3}$ & \tt $~95.28 {\hspace{0.5pt}\times\hspace{0.5pt}} 10^{-3}$ \\
  ~                & proposed  & \tt $~17.89 {\hspace{0.5pt}\times\hspace{0.5pt}} 10^{-3}$ & \tt $~23.53 {\hspace{0.5pt}\times\hspace{0.5pt}} 10^{-3}$ \\ 
  & {\scriptsize (reduction)}  & \tt $~~~~~~72.3\%$                                        & \tt $~~~~~~75.3\%$                                        \\ \midrule
  \multirow{2}*{3} & last seen & \tt $~94.41 {\hspace{0.5pt}\times\hspace{0.5pt}} 10^{-3}$ & \tt $137.41 {\hspace{0.5pt}\times\hspace{0.5pt}} 10^{-3}$ \\
  ~                & proposed  & \tt $~24.03 {\hspace{0.5pt}\times\hspace{0.5pt}} 10^{-3}$ & \tt $~30.25 {\hspace{0.5pt}\times\hspace{0.5pt}} 10^{-3}$ \\ 
  & {\scriptsize (reduction)}  & \tt $~~~~~~74.5\%$                                        & \tt $~~~~~~78.0\%$                                        \\ \midrule
  \multirow{2}*{4} & last seen & \tt $124.70 {\hspace{0.5pt}\times\hspace{0.5pt}} 10^{-3}$ & \tt $179.25 {\hspace{0.5pt}\times\hspace{0.5pt}} 10^{-3}$ \\
  ~                & proposed  & \tt $~32.11 {\hspace{0.5pt}\times\hspace{0.5pt}} 10^{-3}$ & \tt $~33.62 {\hspace{0.5pt}\times\hspace{0.5pt}} 10^{-3}$ \\ 
  & {\scriptsize (reduction)}  & \tt $~~~~~~74.3\%$                                        & \tt $~~~~~~81.2\%$                                        \\ \midrule
  \multirow{2}*{5} & last seen & \tt $163.19 {\hspace{0.5pt}\times\hspace{0.5pt}} 10^{-3}$ & \tt $226.59 {\hspace{0.5pt}\times\hspace{0.5pt}} 10^{-3}$ \\
  ~                & proposed  & \tt $~34.42 {\hspace{0.5pt}\times\hspace{0.5pt}} 10^{-3}$ & \tt $~28.82 {\hspace{0.5pt}\times\hspace{0.5pt}} 10^{-3}$ \\
  & {\scriptsize (reduction)}  & \tt $~~~~~~78.9\%$                                        & \tt $~~~~~~87.3\%$                                        \\ 
  \bottomrule
  \end{tabular}
\end{minipage}
\end{minipage}

~

~

~

\begin{minipage}{1\textwidth}
\begin{minipage}{0.49\textwidth}
\centering
\textbf{HePH dataset}\\
\vspace{0.5ex}
\setlength\tabcolsep{3pt}
\begin{tabular}{cccccccc}
  \toprule
  $h$              & {\bf method} & {\bf vertex error} & {\bf edge error} \\ \midrule
  \multirow{2}*{1} & last seen & \tt $104.93 {\hspace{0.5pt}\times\hspace{0.5pt}} 10^{-3}$ & \tt $134.43 {\hspace{0.5pt}\times\hspace{0.5pt}} 10^{-3}$ \\
  ~                & proposed  & \tt $~39.34 {\hspace{0.5pt}\times\hspace{0.5pt}} 10^{-3}$ & \tt $~47.09 {\hspace{0.5pt}\times\hspace{0.5pt}} 10^{-3}$ \\ 
  & {\scriptsize (reduction)}  & \tt $~~~~~~62.5\%$                                        & \tt $~~~~~~65.0\%$                                        \\ \midrule
  \multirow{2}*{2} & last seen & \tt $194.43 {\hspace{0.5pt}\times\hspace{0.5pt}} 10^{-3}$ & \tt $248.26 {\hspace{0.5pt}\times\hspace{0.5pt}} 10^{-3}$ \\
  ~                & proposed  & \tt $~69.60 {\hspace{0.5pt}\times\hspace{0.5pt}} 10^{-3}$ & \tt $143.24 {\hspace{0.5pt}\times\hspace{0.5pt}} 10^{-3}$ \\ 
  & {\scriptsize (reduction)}  & \tt $~~~~~~64.2\%$                                        & \tt $~~~~~~42.3\%$                                        \\ \midrule
  \multirow{2}*{3} & last seen & \tt $276.44 {\hspace{0.5pt}\times\hspace{0.5pt}} 10^{-3}$ & \tt $349.17 {\hspace{0.5pt}\times\hspace{0.5pt}} 10^{-3}$ \\
  ~                & proposed  & \tt $~99.52 {\hspace{0.5pt}\times\hspace{0.5pt}} 10^{-3}$ & \tt $242.84 {\hspace{0.5pt}\times\hspace{0.5pt}} 10^{-3}$ \\ 
  & {\scriptsize (reduction)}  & \tt $~~~~~~64.0\%$                                        & \tt $~~~~~~30.5\%$                                        \\ \midrule
  \multirow{2}*{4} & last seen & \tt $352.05 {\hspace{0.5pt}\times\hspace{0.5pt}} 10^{-3}$ & \tt $438.32 {\hspace{0.5pt}\times\hspace{0.5pt}} 10^{-3}$ \\
  ~                & proposed  & \tt $114.29 {\hspace{0.5pt}\times\hspace{0.5pt}} 10^{-3}$ & \tt $334.20 {\hspace{0.5pt}\times\hspace{0.5pt}} 10^{-3}$ \\ 
  & {\scriptsize (reduction)}  & \tt $~~~~~~67.5\%$                                        & \tt $~~~~~~23.8\%$                                        \\ \midrule
  \multirow{2}*{5} & last seen & \tt $417.01 {\hspace{0.5pt}\times\hspace{0.5pt}} 10^{-3}$ & \tt $509.83 {\hspace{0.5pt}\times\hspace{0.5pt}} 10^{-3}$ \\
  ~                & proposed  & \tt $140.30 {\hspace{0.5pt}\times\hspace{0.5pt}} 10^{-3}$ & \tt $406.44 {\hspace{0.5pt}\times\hspace{0.5pt}} 10^{-3}$ \\
  & {\scriptsize (reduction)}  & \tt $~~~~~~66.4\%$                                        & \tt $~~~~~~20.3\%$                                        \\ 
  \bottomrule
\end{tabular}
\end{minipage}
\hfill
\begin{minipage}{0.49\textwidth}
\centering
\textbf{Bitcoin dataset}\\
\vspace{0.5ex}
\setlength\tabcolsep{3pt}
\begin{tabular}{cccccccc}
  \toprule
  $h$              & {\bf method} & {\bf vertex error} & {\bf edge error} \\ \midrule
  \multirow{2}*{1} & last seen & \tt $101.60 {\hspace{0.5pt}\times\hspace{0.5pt}} 10^{-3}$ & \tt $112.22 {\hspace{0.5pt}\times\hspace{0.5pt}} 10^{-3}$ \\
  ~                & proposed  & \tt $~29.57 {\hspace{0.5pt}\times\hspace{0.5pt}} 10^{-3}$ & \tt $~30.61 {\hspace{0.5pt}\times\hspace{0.5pt}} 10^{-3}$ \\ 
  & {\scriptsize (reduction)}  & \tt $~~~~~~70.9\%$                                        & \tt $~~~~~~72.7\%$                                        \\ \midrule
  \multirow{2}*{2} & last seen & \tt $198.20 {\hspace{0.5pt}\times\hspace{0.5pt}} 10^{-3}$ & \tt $218.86 {\hspace{0.5pt}\times\hspace{0.5pt}} 10^{-3}$ \\
  ~                & proposed  & \tt $~75.69 {\hspace{0.5pt}\times\hspace{0.5pt}} 10^{-3}$ & \tt $~81.37 {\hspace{0.5pt}\times\hspace{0.5pt}} 10^{-3}$ \\ 
  & {\scriptsize (reduction)}  & \tt $~~~~~~61.8\%$                                        & \tt $~~~~~~62.8\%$                                        \\ \midrule
  \multirow{2}*{3} & last seen & \tt $289.80 {\hspace{0.5pt}\times\hspace{0.5pt}} 10^{-3}$ & \tt $319.52 {\hspace{0.5pt}\times\hspace{0.5pt}} 10^{-3}$ \\
  ~                & proposed  & \tt $135.79 {\hspace{0.5pt}\times\hspace{0.5pt}} 10^{-3}$ & \tt $146.22 {\hspace{0.5pt}\times\hspace{0.5pt}} 10^{-3}$ \\ 
  & {\scriptsize (reduction)}  & \tt $~~~~~~53.1\%$                                        & \tt $~~~~~~54.2\%$                                        \\ \midrule
  \multirow{2}*{4} & last seen & \tt $377.19 {\hspace{0.5pt}\times\hspace{0.5pt}} 10^{-3}$ & \tt $413.35 {\hspace{0.5pt}\times\hspace{0.5pt}} 10^{-3}$ \\
  ~                & proposed  & \tt $193.06 {\hspace{0.5pt}\times\hspace{0.5pt}} 10^{-3}$ & \tt $199.10 {\hspace{0.5pt}\times\hspace{0.5pt}} 10^{-3}$ \\ 
  & {\scriptsize (reduction)}  & \tt $~~~~~~48.8\%$                                        & \tt $~~~~~~51.8\%$                                        \\ \midrule
  \multirow{2}*{5} & last seen & \tt $456.76 {\hspace{0.5pt}\times\hspace{0.5pt}} 10^{-3}$ & \tt $498.31 {\hspace{0.5pt}\times\hspace{0.5pt}} 10^{-3}$ \\
  ~                & proposed  & \tt $240.64 {\hspace{0.5pt}\times\hspace{0.5pt}} 10^{-3}$ & \tt $255.83 {\hspace{0.5pt}\times\hspace{0.5pt}} 10^{-3}$ \\
  & {\scriptsize (reduction)}  & \tt $~~~~~~47.3\%$                                        & \tt $~~~~~~48.7\%$                                        \\ 
  \bottomrule
\end{tabular}
\end{minipage}
\end{minipage}

\end{table}

\clearpage\newpage
\section{Conclusion}
\label{sec:conclusion}

In this work we have proposed a graph prediction approach
comprised of time series modelling combined with an adapted form of Flux Balance Analysis~\cite{Orth2010,Sahu_2021},
a~technique used in biochemistry to reconstruct metabolic networks.
FBA is adapted to incorporate various constraints applicable to unweighted graphs in growing scenarios,
where new vertices and edges appear in consecutive graphs.
The proposed approach addresses problems in previous techniques that have limitations
such as assuming that the number of vertices does not change between consecutive graphs.

Experiments on two synthetic datasets
(constructed via the preferential attachment model~\cite{barabasi1999} with further stochastic behaviour)
and four real datasets
(UCI Message~\cite{Panzarasa2009}, HePH~\cite{Leskovec2007}, Facebook~\cite{viswanath-2009-activity}, Bitcoin~\cite{kumar2016edge})
demonstrate that the proposed approach achieves promising results.
Future avenues of research include extending the proposed approach to weighted and directed graphs,
as well as using more sophisticated graph similarity measures via exploiting spectral geometry~\cite{ElGhawalby_2008,Liu_2025}.

~

\begin{small}
\noindent
\textbf{Acknowledgements}.
We  would like to thank our colleague Dan Pagendam (Data61/CSIRO) for discussions leading to improvements of this work.
Sevvandi Kandanaarachchi is part of the Australian Research Council (ARC)
Industrial Transformation Training Centre in Optimisation Technologies, Integrated Methodologies, and Applications (OPTIMA),
Project ID IC200100009.
\end{small}

\def~{\,}  % refine ~ character to be a shorter non-breaking space; https://texfaq.org/FAQ-activechars

\bibliographystyle{ieee_mod}  %% for arxiv
%bibliographystyle{splncs04}
\bibliography{references}

\begin{thebibliography}{10}\interlinepenalty=10000\itemsep=0.5ex

\bibitem{Aiello_2012}
L.~M. Aiello, A.~Barrat, R.~Schifanella, C.~Cattuto, B.~Markines, and
  F.~Menczer.
\newblock Friendship prediction and homophily in social media.
\newblock {\em ACM Transactions on the Web}, 6(2):1--33, 2012.

\bibitem{barabasi1999}
A.-L. Barab{\'{a}}si and R.~Albert.
\newblock {Emergence of Scaling in Random Networks}.
\newblock {\em Science}, 286(5439):509--512, 1999.

\bibitem{Arrieta_2020}
A.~Barredo~Arrieta et~al.
\newblock Explainable artificial intelligence ({XAI}): Concepts, taxonomies,
  opportunities and challenges toward responsible~{AI}.
\newblock {\em Information Fusion}, 58:82--115, 2020.

\bibitem{Baumann_2014}
A.~Baumann, B.~Fabian, and M.~Lischke.
\newblock Exploring the {Bitcoin} network.
\newblock In {\em International Conference on Web Information Systems and
  Technologies}, volume~2, pages 369--374, 2014.

\bibitem{Myttenaere_2016}
A.~de~Myttenaere, B.~Golden, B.~Le~Grand, and F.~Rossi.
\newblock Mean absolute percentage error for regression models.
\newblock {\em Neurocomputing}, 192:38--48, 2016.

\bibitem{Ekle_2024}
O.~A. Ekle and W.~Eberle.
\newblock Anomaly detection in dynamic graphs: A comprehensive survey.
\newblock {\em ACM Transactions on Knowledge Discovery from Data}, 18(8):1--44,
  2024.

\bibitem{ElGhawalby_2008}
H.~ElGhawalby and E.~R. Hancock.
\newblock Measuring graph similarity using spectral geometry.
\newblock {\em Lecture Notes in Computer Science}, 5112:517--526, 2008.

\bibitem{graphiso}
M.~Grohe and P.~Schweitzer.
\newblock The graph isomorphism problem.
\newblock {\em Communications of the ACM}, 63(11):128--134, 2020.

\bibitem{HyndmanBook2013}
R.~J. Hyndman and G.~Athanasopoulos.
\newblock {\em {Forecasting: Principles and Practice}}.
\newblock OTexts, 3rd edition, 2021.

\bibitem{Kandanaarachchi_2024}
S.~Kandanaarachchi, C.~Sanderson, and R.~J. Hyndman.
\newblock Extreme value modelling of feature residuals for anomaly detection in
  dynamic graphs.
\newblock In {\em Int.~Conf.~Soft Computing \& Machine Intelligence}, page
  32–37, 2024.

\bibitem{Kazemi2020}
S.~M. Kazemi, R.~Goel, K.~Jain, I.~Kobyzev, A.~Sethi, P.~Forsyth, and
  P.~Poupart.
\newblock Representation learning for dynamic graphs: A survey.
\newblock {\em Journal of Machine Learning Research}, 21(70):1--73, 2020.

\bibitem{Khanam_2023}
K.~Z. Khanam, G.~Srivastava, and V.~Mago.
\newblock The homophily principle in social network analysis: A survey.
\newblock {\em Multimedia Tools and Applications}, 82(6):8811--8854, 2023.

\bibitem{Kim_2017}
K.~Kim and J.~Altmann.
\newblock Effect of homophily on network formation.
\newblock {\em Communications in Nonlinear Science and Numerical Simulation},
  44:482--494, 2017.

\bibitem{KIM2024971}
K.~Kim and H.-S. Oh.
\newblock Network time series forecasting using spectral graph wavelet
  transform.
\newblock {\em International Journal of Forecasting}, 40(3):971--984, 2024.

\bibitem{Kumar2020}
A.~Kumar, S.~S. Singh, K.~Singh, and B.~Biswas.
\newblock {Link prediction techniques, applications, and performance: A
  survey}.
\newblock {\em Physica A: Statistical Mechanics and its Applications}, 553,
  2020.

\bibitem{kumar2016edge}
S.~Kumar, F.~Spezzano, V.~Subrahmanian, and C.~Faloutsos.
\newblock Edge weight prediction in weighted signed networks.
\newblock In {\em IEEE International Conference on Data Mining (ICDM)}, pages
  221--230, 2016.

\bibitem{Leskovec2007}
J.~Leskovec, J.~Kleinberg, and C.~Faloutsos.
\newblock {Graph evolution: Densification and shrinking diameters}.
\newblock {\em ACM Transactions on Knowledge Discovery from Data}, 1(1), 2007.

\bibitem{Liu_2025}
Z.~Liu, N.~Liu, Y.~Chen, Z.~Wen, J.~He, and D.~Li.
\newblock Graph theory-based deep graph similarity learning: A unified survey
  of pipeline, techniques, and challenges.
\newblock {\em Transactions on Machine Learning Research}, 2025.

\bibitem{ma2021deep}
G.~Ma, N.~K. Ahmed, T.~L. Willke, and P.~S. Yu.
\newblock Deep graph similarity learning: A survey.
\newblock {\em Data Mining and Knowledge Discovery}, 35:688--725, 2021.

\bibitem{Orth2010}
J.~D. Orth, I.~Thiele, and B.~{\O}. Palsson.
\newblock What is flux balance analysis?
\newblock {\em Nature Biotechnology}, 28(3):245--248, 2010.

\bibitem{Panzarasa2009}
P.~Panzarasa, T.~Opsahl, and K.~M. Carley.
\newblock {Patterns and dynamics of users' behavior and interaction: Network
  analysis of an online community}.
\newblock {\em Journal of the American Society for Information Science and
  Technology}, 60(5):911--932, 2009.

\bibitem{Sahu_2021}
A.~Sahu, M.-A. Bl\"{a}tke, J.~J. Szyma\'{n}ski, and N.~T\"{o}pfer.
\newblock Advances in flux balance analysis by integrating machine learning and
  mechanism-based models.
\newblock {\em Computational and Structural Biotechnology Journal},
  19:4626--4640, 2021.

\bibitem{Sanderson_2023}
C.~Sanderson, D.~Douglas, and Q.~Lu.
\newblock Implementing responsible {AI}: Tensions and trade-offs between ethics
  aspects.
\newblock In {\em International Joint Conference on Neural Networks}, 2023.

\bibitem{Souid_2024}
H.~E. Souid, L.~Ody, Y.~Achenchabe, V.~Lemaire, G.~Aversano, and S.~Skhiri.
\newblock Temporal graph generative models: An empirical study.
\newblock In {\em Workshop on Machine Learning and Systems ({EuroMLSys})}, page
  18–27, 2024.

\bibitem{viswanath-2009-activity}
B.~Viswanath, A.~Mislove, M.~Cha, and K.~P. Gummadi.
\newblock On the evolution of user interaction in {Facebook}.
\newblock In {\em ACM SIGCOMM Workshop on Social Networks}, 2009.

\end{thebibliography}

\end{document}